# Contrast Enhancement or Noise Reduction? On Improving Cervical Cancer Classification

**Ach Khozaimi[1*], Ulfatun Nahdhiyah[2]**
[1]Department of Computer Science, University of Trunodjoyo Madura, Indonesia
[2]Anatomical Pathology Laboratory, District General Hospital (RSUD Syarifah Ambami Rato Ebu), Indonesia

**Abstract.**
**Purpose:** Cervical cancer is one of the leading causes of mortality worldwide. Deep learning has shown promising performance in medical image classification. The influence of image preprocessing algorithms on classification performance remains insufficiently investigated in the literature. This research aims to evaluate the impact of image preprocessing algorithms on the performance of CNNs for Pap smear image classification.
**Methods:** Three CNN architectures (ResNet-34, MobileNet-V2, and DenseNet-121) were trained and evaluated using the SIPaKMeD dataset. Two preprocessing algorithms were applied: the PMD filter for noise reduction and CLAHE for contrast enhancement. The model performance was assessed using a confusion matrix.
**Result:** Preprocessing improved the classification performance of all models. CLAHE significantly increased the accuracy of ResNet-34 from 76.73% to 84.16% and DenseNet-121 from 76.73% to 84.16%. The PMD filter yielded limited improvement and slightly reduced the MobileNet-V2 performance.
**Novelty:** This research provides a systematic comparison of contrast enhancement and noise reduction techniques across CNN architectures. This research demonstrates that contrast enhancement is more effective than noise reduction in improving CNN performance. The research provides new pipelines for improving cervical cancer classification.

## INTRODUCTION

Cervical cancer represents a substantial public health concern. It is the fourth most prevalent cancer among women worldwide [1]. Cervical cancer mainly results from a persistent infection with high-risk human papillomavirus (HPV) types [2]. Screening tests such as the Papanicolaou (Pap) smear test have proven effective in detecting precancerous cellular abnormalities of cervical cancer [3]. Manual evaluation of Pap smear slides is inherently subjective and heavily relies on the expertise of cytotechnologists and pathologists. It makes them susceptible to diagnostic variability and human error [4]. The demand for automated screening systems is rising as the need for large-scale, reliable screening grows. It can effectively and efficiently assist or even replace manual cytological evaluations [5].

Convolutional neural networks (CNNs) have driven a major paradigm shift in medical analysis. These networks have facilitated automated, data-driven feature extraction and classification [2]. CNNs have been successfully used in medical image classification [6]. CNNs outperform traditional machine learning techniques [7]. ResNet-34 and DenseNet-121 architectures have gained prominence for their deep hierarchical structures and ability to address common training challenges [8][9]. ResNet-34 employs residual learning to make training deeper networks easier. ResNet-34 enables more efficient gradient flow [8]. The ResNet-34 architecture enhances the model's accuracy, sensitivity, and specificity when processing the dataset [10]. DenseNet-121 establishes direct connections between all layers, thereby facilitating feature reuse and mitigating vanishing gradient issues. The DenseNet-121 model achieves the highest accuracy in COVID-19 image classification [9]. These models are especially effective for classifying medical images. MobileNet-V2 is a lightweight CNN architecture. MobileNet-V2 also performs well in medical image classification [11].

*Corresponding author.
Email addresses: khozaimi@trunojoyo.ac.id (Khozaimi)

Recent studies show that deep learning effectively classifies cervical cancer from Pap smear images. Numerous researchers have reported promising outcomes using transfer learning [12], advanced CNN architectures [13] and hybrid deep learning frameworks [14]. Studies have primarily focused on enhancing network architectures, attention mechanisms [15], feature fusion strategies [16], and ensemble learning techniques [17] to improve classification accuracy. Most previous research has focused on model development rather than on examining the impact of image quality enhancement. The role of preprocessing techniques in CNN-based cervical cancer classification remains inadequately explored.

The effectiveness of CNNs depends significantly on the image quality [18]. Pap smear images often display significant variability in staining, lighting, and slide preparation, leading to artifacts such as poor contrast and image noise [19]. These issues can obscure critical features of cervical cells. These features are essential for the accurate classification of cervical cells [20]. Therefore, preprocessing steps designed to enhance image quality are crucial to maximizing deep learning performance. Noise reduction and contrast enhancement are two algorithms to improve visual clarity and structural integrity in cytological images. This research assesses the performance of the Perona-Malik Diffusion (PMD) filter and Contrast Limited Adaptive Histogram Equalization (CLAHE). The PMD filter performs anisotropic diffusion, reducing noise while preserving edges and important cellular boundaries [21]. CLAHE improves local contrast by using "histogram equalization" to adjust small images' regions [22].

Although prior studies show CNN effectiveness, several research gaps remain. First, most studies focus on developing more advanced classification models, placing less emphasis on the importance of image preprocessing. Second, existing studies typically evaluate either contrast enhancement or noise reduction individually. Third, comparative investigations involving multiple CNN architectures under identical preprocessing conditions remain scarce. It remains unclear whether contrast enhancement or noise reduction yields greater benefits for cervical cancer classification. This research systematically investigates the impact of CLAHE and PMD filter preprocessing techniques on the performance of three representative CNN architectures (ResNet-34, MobileNet-V2, and DenseNet-121). This research utilized the SIPaKMeD dataset. This research seeks to thoroughly examine the impact of image quality enhancement on CNN performance. The proposed framework is evaluated using a confusion matrix under three experimental conditions: original Pap smear images, CLAHE-enhanced images, and PMD-filtered images. The simulation results show that CLAHE consistently outperforms PMD and the original images. CLAHE achieved the highest classification accuracies of 84.16%, 83.17%, and 84.65% for ResNet-34, DenseNet-121, and MobileNet-V2, respectively. These findings indicate that contrast enhancement outperforms noise reduction in improving CNN accuracy for cervical cancer classification. The study provides new pipelines for improving cervical cancer classification.

## METHODS

### Proposed methods

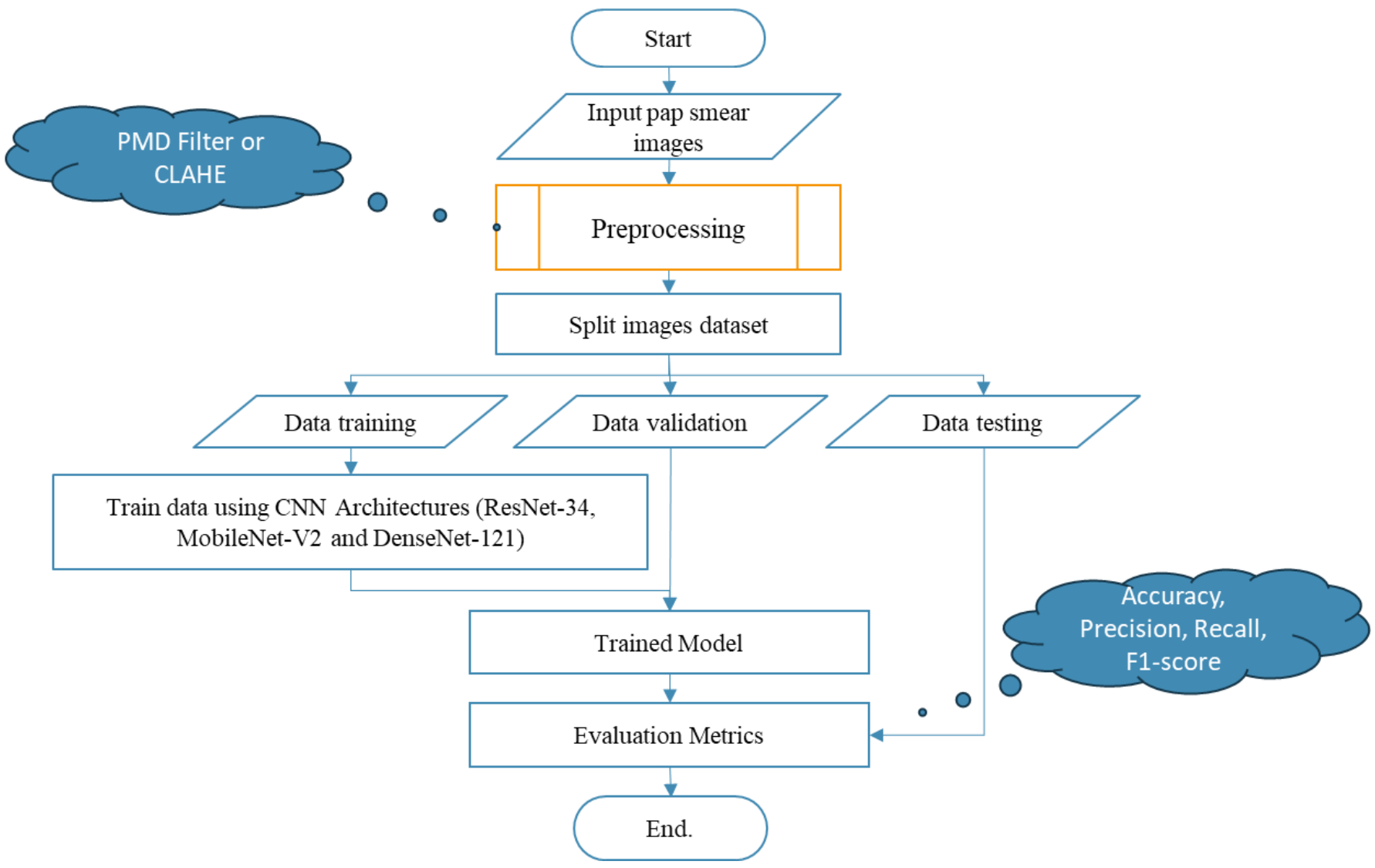

Figure 1. Research Workflow of the Proposed Cervical Cancer Classification Framework

Figure 1 illustrates the workflow for cervical cell image classification using CNN architectures with image preprocessing algorithms. The process starts with input images from the SIPaKMeD dataset. These images are enhanced using either CLAHE to improve local contrast or a PMD filter to reduce noise and preserve structural edges. The detailed explanations of the CLAHE and PMD filtering methods are provided in Sections 2.3 and 2.4, respectively. These preprocessed images are then fed into CNN architectures (ResNet-34, MobileNet-V2, and DenseNet-121). These CNN architectures are trained to classify cervical cell types based on learned feature representations. The performance of CNN architectures is assessed using a confusion matrix. The results are then compared across preprocessing methods to assess their impact. The specifics of the setting parameters, hyperparameters, and the hardware and software used are delineated in Section 2.6. This section also presents the flowchart for the CLAHE and PMD filter algorithms.

**SIPaKMeD dataset**

This research utilized the SIPaKMeD dataset. This dataset is one of the most frequently adopted benchmark datasets for cervical cytology image analysis. Developed to support automated cervical cancer screening research, it provides a diverse collection of Pap smear cell images accompanied by reliable expert annotations. The dataset has been extensively used to evaluate the performance of deep learning models for cervical cell classification. It contains 4,049 segmented and cropped single-cell images derived from 966 original multi-cell microscopic images. The images were captured using a CCD camera under 400× magnification [23]. Although the dataset does not represent all possible cervical cell variations, each image is carefully categorized into five distinct cell classes based on cytopathologist annotations:

1. Superficial-Intermediary (SI) is normal squamous epithelial cells,
2. Parabasal (PB) is immature squamous cells seen in postmenopausal women or in atrophic smears,
3. Koilocytotic (K) is indicative of HPV infection and possible precancerous changes,
4. Dyskeratotic (D) refers to cells that exhibit abnormal keratinization, frequently observed in cases of dysplasia or carcinoma,
5. Metaplastic (M) is when cells transform into squamous and glandular types.

Tabel 1. Five sample images from five classes of cervical cancer

| No. | SI | PB | K | D | M |
|---|---|---|---|---|---|
| 1 | 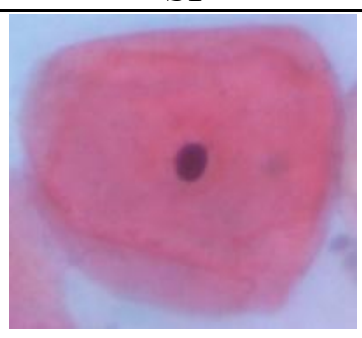 | 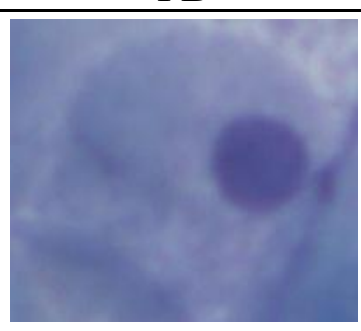 | 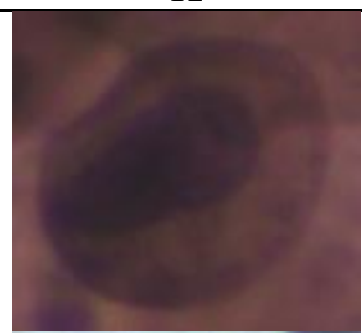 | 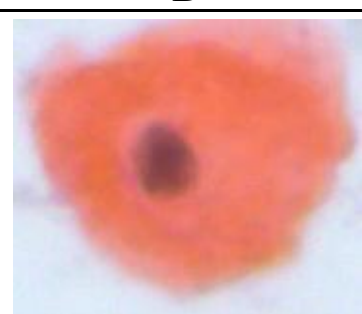 | 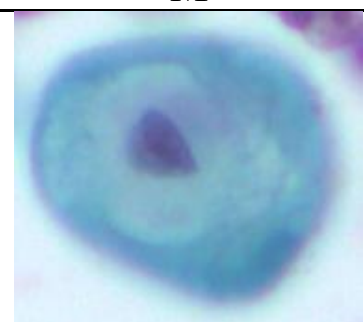 |
| 2 | 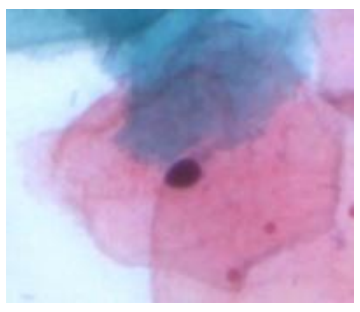 | 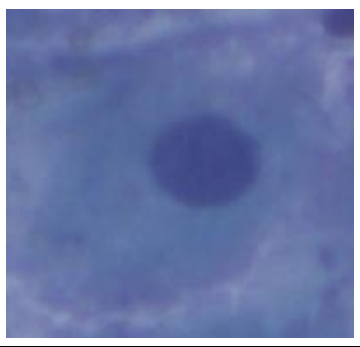 | 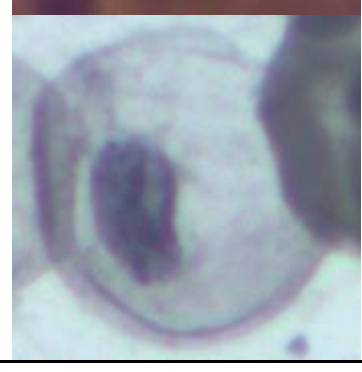 | 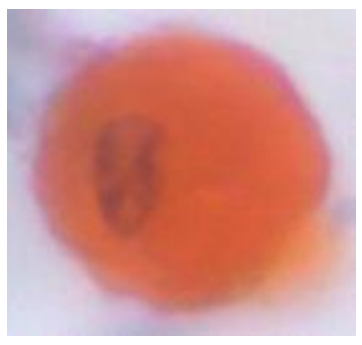 | 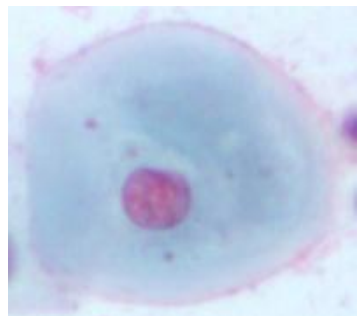 |

Each cell image in the dataset is provided in RGB format with a resolution of 256 × 256 pixels. The dataset has approximately equal numbers of samples per class. It facilitates model training and evaluation without the need for complex resampling strategies. The images exhibit variability in staining intensity, cellular overlap, and background noise. SIPaKMeD a challenging yet realistic dataset for evaluating the performance of image classification models under real-world conditions. This dataset challenges the development of robust cervical cancer screening models [24]. The SIPaKMeD dataset is openly available at https://bit.ly/SIPaKMeD. Table 1 presents five Pap smear cell samples (rows 1–5) across different staining or imaging conditions labeled as SI, PB, K, D, and M. Each cell shows typical cervical morphology but suffers from visible degradation in image quality. The images show noticeable noise and low contrast, obscuring nuclear and cytoplasmic boundaries. These conditions make visual inspection and automated analysis more challenging.

**Contrast enhancement**

Image contrast enhancement is a crucial preprocessing step in medical image analysis, particularly when working with microscopic images such as Pap smear slides. Contrast defines the visual difference between objects in an image, such as the nucleus and cytoplasm of a cell, and insufficient contrast can obscure essential morphological features required for classification. Enhancing contrast reveals subtle details and boundaries, thereby facilitating improved feature extraction for both manual inspection and automated classification using deep learning[25].

Histogram Equalization (HE) is among the most widely used techniques for improving image contrast. It adjusts the pixel intensities of an image to span the full available dynamic range, thereby flattening and broadening the histogram [26]. However, standard HE can lead to over-enhancement and noise amplification, especially in medical images with uniform backgrounds or delicate textures. CLAHE was developed to address these limitations as an enhancement over traditional HE [27]. CLAHE works on small regions (tiles) instead of the whole image. CLAHE applies HE to each tile separately. The outcomes are then blended using bilinear interpolation to smooth out any artificial boundary lines. CLAHE incorporates a contrast-limited mechanism to prevent noise amplification by clipping the histogram [28]. CLAHE algorithm and formulas are as follows. Let $I(x, y)$ be the grayscale input image.

1. Image Tiling
   The image is segmented into 8×8 or 16×16 pixels, denoted as $T_{ij}$, where $i, j$ index the tiles.

2. Histogram Calculation
   For each tile $T_{ij}$, compute the histogram $H_{ij}(r_k)$ of pixel intensities $r_k \in [0, L-1]$, where $L$ is the number of possible intensity levels.

3. Histogram Clipping
   A clip limit $C$ is defined as a threshold for the maximum height of the histogram bins. Any bin exceeding $C$ is clipped, and the surplus is shared across all histogram bins:

$$H'_{ij}(r_k) = \min\big(H_{ij}(r_k), C\big) \tag{1}$$

   The clipped pixels are redistributed uniformly:

$$\Delta = \frac{1}{L}\sum_{r_k} \max\big(H_{ij}(r_k) - C, 0\big) \tag{2}$$

$$H''_{ij}(r_k) = H'_{ij}(r_k) + \Delta \tag{3}$$

4. CDF Calculation and Mapping
   Compute the cumulative distribution function (CDF) for each clipped histogram:

$$\mathrm{CDF}_{ij}(r_k) = \sum_{l=0}^{k} \frac{H''_{ij}(r_l)}{n} \tag{4}$$

   where $n$ is the total number of pixels in the tile. The new pixel intensity is then mapped as:

$$I_{\mathrm{enh}}(x, y) = (L-1) \cdot \mathrm{CDF}_{ij}\big(I(x, y)\big) \tag{5}$$

5. Bilinear Interpolation
   Bilinear interpolation is applied between adjacent tiles to avoid discontinuities at tile borders.

**Noise reduction**

Reducing noise in images is crucial for medical analysis, particularly in Pap smear images. Pap smear images often contain noise from sample preparation, staining, and image capture processes. Noise can obscure subtle anatomical structures and lead to misclassification in machine learning models. Traditional linear smoothing filters, such as Gaussian or mean filters, blur important edge information and noise, making them undesirable in medical diagnostics. The PMD filter was introduced as an edge-preserving method for noise reduction. The PMD filter evolves the image over time using a partial differential equation (PDE). It selectively smooths areas based on the gradient magnitude, reducing image noise in homogeneous regions while preserving edges [29]. The PMD process is described by the subsequent equation:

$$\frac{\partial I(x,y,t)}{\partial t} = \nabla \cdot [c(|\nabla I|) \cdot \nabla I] \tag{6}$$

Here, $I(x, y, t)$ represents the image intensity at a spatial position $(x, y)$ and time $t$, $\nabla I$ is the image gradient, and $c(|\nabla I|)$ The diffusion coefficient determines how quickly diffusion occurs, depending on the gradient's magnitude [30]. Two common choices for the diffusion coefficient function are:

1. Inverse function

$$c(|\nabla I|) = \frac{1}{1+\left(\frac{|\nabla I|}{K}\right)^2} \tag{7}$$

2. Exponential function (edge-stopping function)

$$c(|\nabla I|) = \exp\left(-\left(\frac{|\nabla I|}{K}\right)^2\right) \tag{8}$$

where $K$ is a contrast threshold parameter for edge sensitivity. Smaller values of $K$ preserve finer edges, while larger values lead to more aggressive smoothing. The PMD filter evolves the image through multiple iterations (or time steps), each reducing noise in flat regions while maintaining or enhancing the contrast around edges. This approach is especially effective for medical images, where maintaining clear boundaries is essential [31].

**CNN architectures**

CNNs are deep learning models used for image classification. CNNs extract hierarchical features from images through convolutional layers. It enables the identification of both simple visual patterns and complex structures [32]. This research uses three CNN models (ResNet-34, MobileNet-V2, and DenseNet-121) to assess the effects of image preprocessing methods on cervical cancer classification. These architectures were selected because they represent different network designs and computational complexities. Table 2 summarizes the characteristics of the ResNet-34, MobileNet-V2, and DenseNet-121 architectures used in this research.

Table 2. Characteristics of ResNet-34, MobileNet-V2, and DenseNet-121 architectures

| Model | Characteristics | Layers | Parameters | Ref |
|---|---|---|---|---|
| ResNet-34 | Employs residual (skip) connections to enhance gradient propagation and alleviate the vanishing gradient problem. | 34 | ~21.8 M | [33] |
| DenseNet-121 | Employs dense connectivity, in which each layer receives feature maps from all preceding layers, thereby enhancing feature reuse and improving information flow. | 121 | ~8.0 M | [9] |
| MobileNet-V2 | It uses depthwise separable convolutions and inverted residual blocks to boost speed without losing performance. | 53 | ~3.5 M | [34] |

ResNet-34 represents a deep residual learning approach. DenseNet-121 employs dense feature connectivity to improve feature reuse. MobileNet-V2 provides a lightweight alternative through depth-wise separable convolutions. The diversity of these architectures enables a comprehensive evaluation of preprocessing effects across different network designs and computational complexities. All models started with ImageNet pre-trained weights and were trained using identical experimental settings.

**Simulation setting**

This study assessed the impact of two preprocessing methods: CLAHE and the PMD filter. Figure 2 illustrates the preprocessing pipeline. It involves decomposing each image into its RGB channels. CLAHE or the PMD filter is applied separately to each channel to improve contrast and suppress noise. After enhancement, the R, G, and B components are recombined to generate the final processed image.

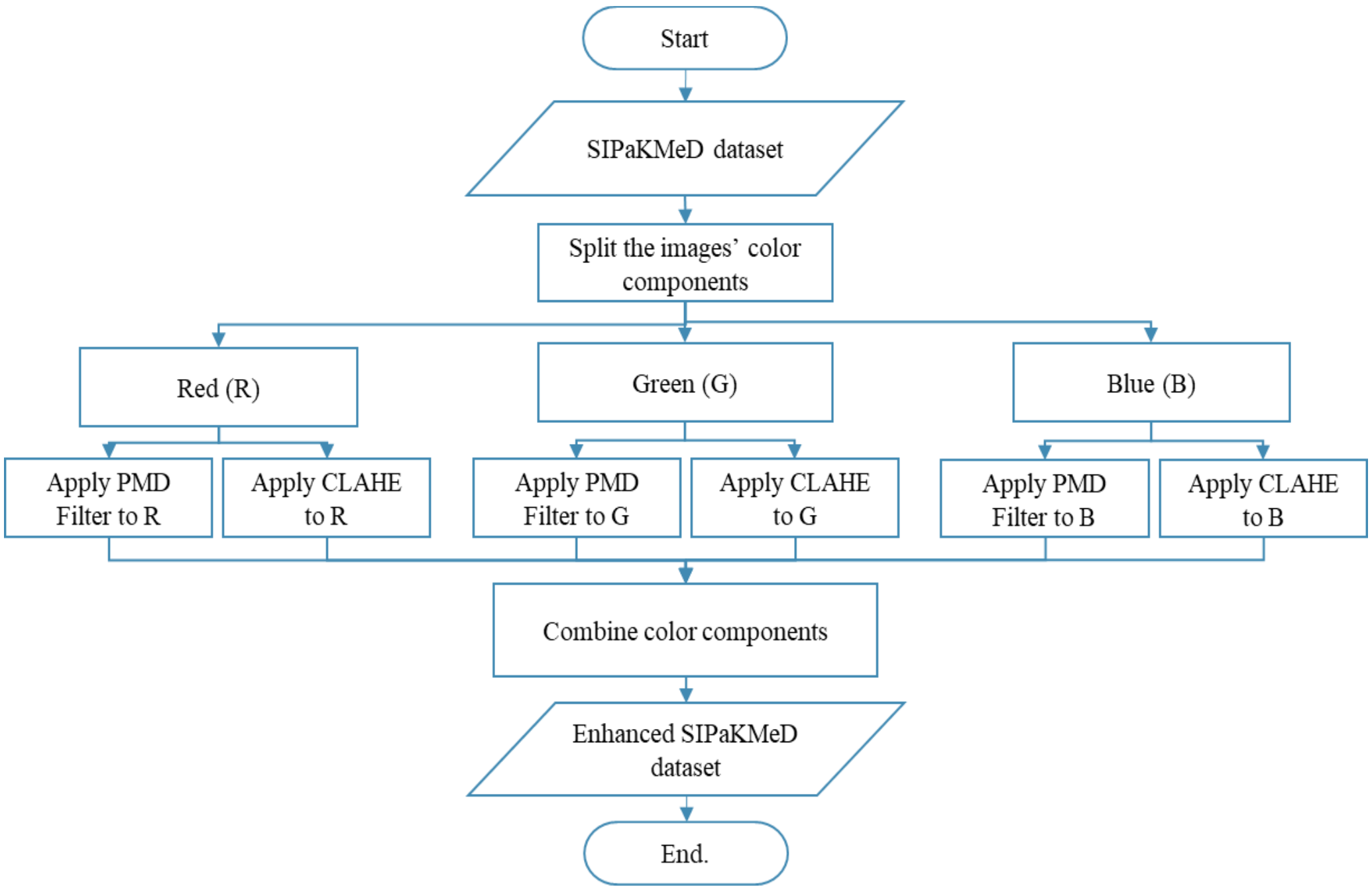

Figure 2. Image preprocessing step with PMD Filter or CLAHE

The PMD filter was employed to suppress image noise while maintaining important edge information, which is essential for preserving the morphological characteristics of cervical cells. Unlike conventional smoothing techniques that often blur object boundaries, PMD selectively diffuses within homogeneous regions and limits diffusion near edges. However, its effectiveness depends heavily on the choice of parameters. The filter is controlled by three parameters: the iterations (N), the time step (λ), and the diffusion coefficient (κ). Based on recommendations from previous studies, the values were fixed at N = 20, λ = 0.25, and κ = 20 [35].

CLAHE was applied to improve image contrast. This algorithm enhances local image details by adjusting the intensity distribution within small regions of the image. While CLAHE can reveal subtle cellular structures that may not be visible in the original images, excessive enhancement may also amplify unwanted artifacts and noise. Its performance is mainly influenced by the clip limit and block size parameters. Lower clip-limit values tend to prevent over-enhancement, whereas higher values produce stronger contrast enhancement at the risk of introducing noise. Likewise, smaller blocks emphasize local details, while larger blocks provide smoother intensity transitions. The CLAHE parameters were set to *block_size* = 8 × 8 and *clip_limit* = 2.0 [36]. The preprocessing was applied independently to each RGB channel, allowing effective enhancement while preserving the original color characteristics of cervical cell images.

The dataset was split into training (80%), testing (10%), and validation (10%) sets. All images were resized to 224 × 224 pixels. Three pre-trained models (ResNet34, DenseNet121, and MobileNet-V2) were initialized with ImageNet weights. The training process was conducted for 30 epochs using the Adamax optimizer with a learning rate of 0.001 and a batch size of 32 [4]. Cross-entropy loss was adopted as the optimization objective for the multi-class classification. To ensure that performance differences originated from the preprocessing strategies and network architectures rather than training settings, all models were trained under identical experimental conditions.

All simulations were performed on a Windows 11 workstation equipped with an AMD Ryzen 5 5500 processor, a NVIDIA GeForce RTX 3060 graphics card with 12 GB of VRAM, and 32 GB of RAM. The implementation was developed using Python 3.11, PyTorch 2.6, TorchVision 0.21, and OpenCV 4.9. In addition, a fixed random seed value of 42 was used throughout the experiments to improve reproducibility and reduce variation due to random initialization.

**Performance metrics**

A confusion matrix was employed to assess the performance of the CNN models in this research. A confusion matrix demonstrates the degree to which predicted classes correspond with actual classes. It contains four key values, and we derived four essential performance metrics:

$$\text{Accuracy} = \frac{TP+TN}{TP+TN+FP+FN} \quad (9)$$

$$\text{Recall} = \frac{TP}{TP+FN} \quad (10)$$

$$\text{Precision} = \frac{TP}{TP+FP} \quad (11)$$

$$\text{F1-score} = 2 \cdot \frac{\text{Precision}\cdot\text{Recall}}{\text{Precision}+\text{Recall}} \quad (12)$$

We calculated these metrics for each class in the multi-class setting and then averaged the results. These measures show the model's overall performance and if it struggles with certain cell types. The confusion matrix also helps identify which classes are frequently confused with others, which is beneficial for improving model accuracy in future studies [37].

# RESULT AND DISCUSSION

## Result

### *Qualitative visual inspection*

Figure 3 presents a visual comparison of the original Pap smear images and those enhanced using two preprocessing techniques: CLAHE and PMD filter. The original images, presented in column (a), display relatively low contrast and subtle noise, which can obscure critical cellular features. While the nucleus and cytoplasm are present, they are indistinguishable, making it more challenging to extract meaningful structural information, especially at their boundaries.

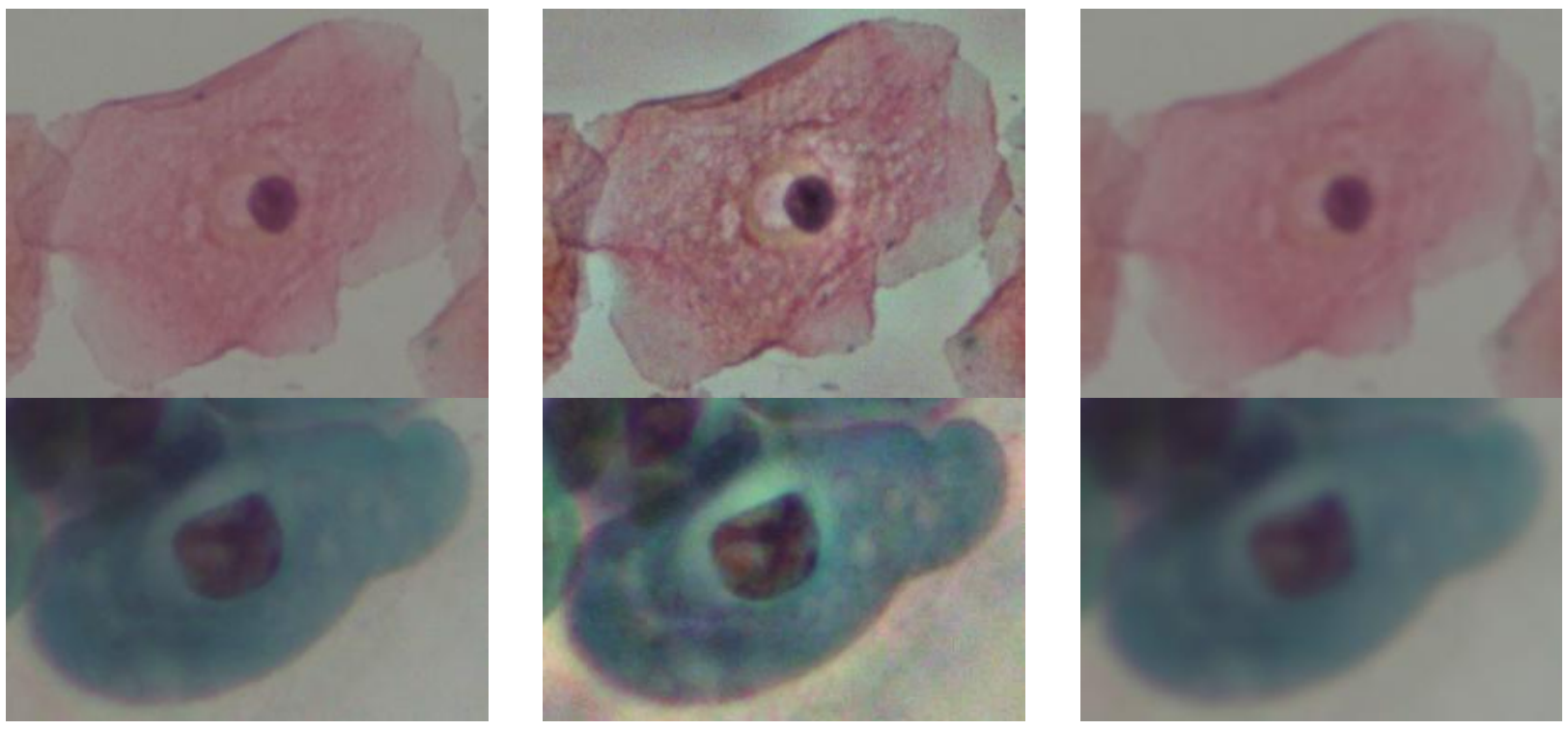

a. b. c.

Figure 3. a. original pap smear images, b. Enhanced pap smear images with CLAHE, c. enhanced pap smear images with PMD filter

Images enhanced with CLAHE (b) reveal a significant improvement in local contrast. The nuclear boundaries become sharper, and the overall brightness distribution is more balanced. Fine details within both the nucleus and the cytoplasm are enhanced, making cellular structures easier to interpret. This improved visibility supports more effective feature extraction by deep learning models, enabling differentiation among cell types based on morphological cues. In comparison, the images processed with the PMD filter (Figure 3.c) appear smoother and show reduced noise, particularly in the background and cytoplasmic regions. The PMD filter effectively suppresses unwanted variations while preserving the general shape of the cells. However, some fine textures and edges appear softened. It potentially diminishes the clarity of certain cellular features. The PMD filter helps suppress noise and preserve structure. It does not improve contrast as effectively as CLAHE. The CLAHE enhances morphological features and structural clarity more effectively. PMD is a complementary technique for noise reduction. It does not sufficiently improve the visual distinctiveness required for accurate classification. These observations support the use of contrast enhancement as a critical preprocessing step in automated Pap smear image analysis.

***ResNet-34 performances***
The experimental results demonstrate the significant impact of image preprocessing on ResNet34's classification performance when applied to Pap smear images. Without any preprocessing, the model achieved an accuracy of 76.73%, a recall of 76.90%, an F1-score of 76.19%, and a precision of 79.12%, establishing the baseline performance. These values indicate a moderately effective model but suggest room for improvement through better feature representation. The ResNet-34 performance with respect to preprocessing is shown in Figure 4.

Applying CLAHE as a preprocessing step yielded a substantial improvement across all performance metrics. This performance boost indicates that CLAHE effectively enhances the contrast and visibility of critical features, such as nuclei and cytoplasmic boundaries, in Pap smear images. It enables ResNet-34 to learn more discriminative representations. CLAHE can improve local contrast without amplifying noise. This appears advantageous for medical imaging, where subtle variations in cellular morphology are crucial for precise classification.

The PMD filter also provided noticeable improvements over the original dataset, although not as substantial as those achieved by CLAHE. The PMD-preprocessed images attained an accuracy of 79.21%, a precision of 80.49%, a recall of 79.39%, and an F1-score of 78.76%. The PMD helps reduce noise and enhance structural integrity by preserving important edges. Its effect on contrast enhancement is less pronounced than CLAHE's. The PMD-preprocessed images still present challenges in distinguishing overlapping or weakly contrasted cellular components. The comparative results highlight the significance of preprocessing in pap smear image analysis. CLAHE stands out as a highly effective technique for improving Pap smear image quality, followed by the PMD filter.

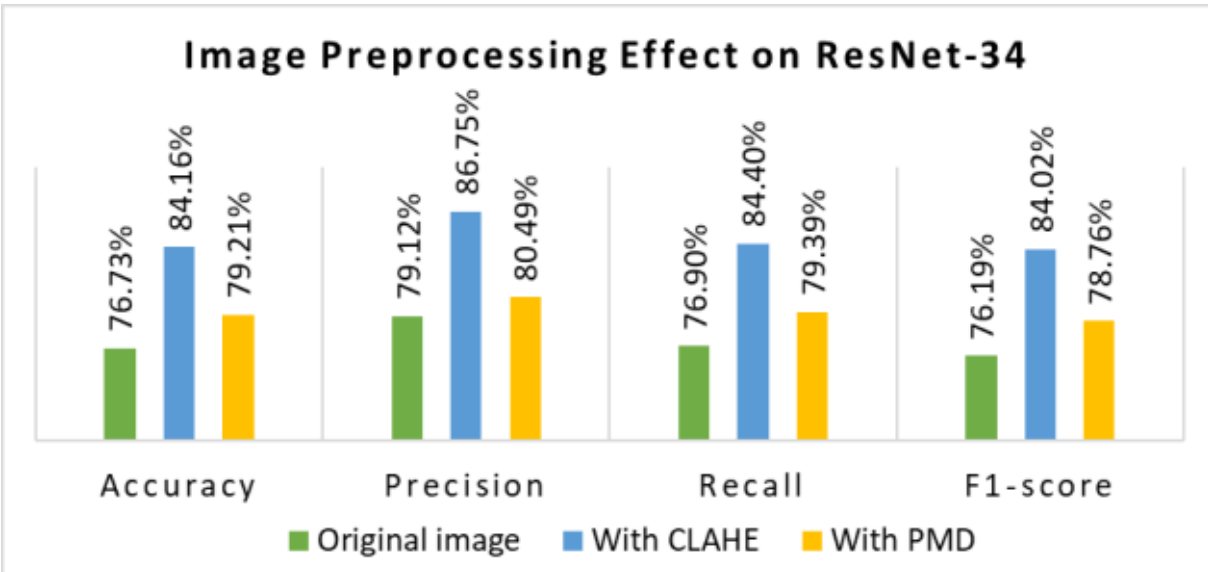

Figure 4. Effect of image preprocessing on ResNet-34 performance

***MobileNet-V2 performance***
The experimental results demonstrate that image preprocessing has a measurable yet moderate impact on the analysis performance of MobileNet-V2 on Pap smear images. The model achieved a strong baseline performance, with an accuracy of 83.91% using the original images. It indicates that MobileNet-V2's inherent efficiency in extracting features from medical cytological images is evident, even without prior enhancement. Precision, F1-score, and recall also demonstrated strong, consistent classification performance. It indicates that the model is appropriate for efficient yet accurate image analysis. The preprocessing effect is shown in Figure 5.

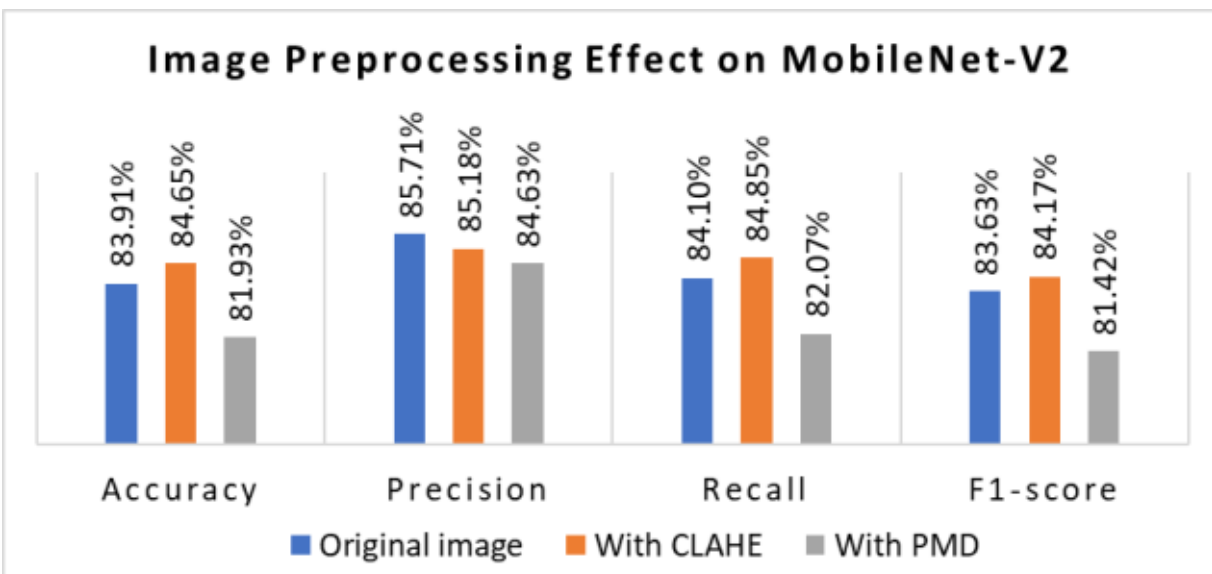

Figure 5. Effect of image preprocessing on MobileNet-V2 performance

When CLAHE was applied as a preprocessing technique, a slight improvement was observed across all performance metrics. Accuracy increased to 84.65%, and the F1-score rose to 84.17%. Image enhancement using CLAHE suggests that it effectively enhances local contrast and improves the model's ability to differentiate subtle textural patterns and morphological features within cervical cells. While the performance gain was not as dramatic as observed in deeper architectures. The enhancement indicates that even compact models like MobileNet-V2 benefit from localized contrast enhancement, especially in tasks involving fine structural analysis. Applying the PMD filter resulted in a slight performance degradation, with accuracy dropping to 81.93%. While PMD effectively reduces background noise and preserves general structural boundaries, it also inadvertently suppresses subtle high-frequency features critical for classification. This slight loss in detail appears to affect MobileNet-V2 more noticeably, likely due to its depthwise separable convolutions and lighter parameterization. It is more sensitive to feature preservation in image preprocessing stages.

The MobileNet-V2 performs well even on original images. Applying CLAHE to Pap smear images yields modest yet consistent improvements, reinforcing the importance of contrast enhancement for cervical cell image classification. Applying a PMD filter for Pap smear images may compromise the feature richness required by lightweight architectures. These findings suggest that contrast enhancement techniques are generally more effective than noise reduction in optimizing MobileNet-V2 performance for cervical cancer classification.

***DenseNet-121 performances***

The experimental evaluation of DenseNet-121 on Pap smear images highlights the crucial role of image preprocessing in improving classification performance, as shown in Figure 6. On the original images, DenseNet-121 achieved 76.73% accuracy, 79.81% precision, 76.85% recall, and 76.04% F1-score. This baseline performance indicates moderate model efficacy but also reveals suboptimal image features or noise that may impede effective feature extraction and classification.

Applying CLAHE significantly boosted the model's performance. Accuracy increased to 83.17%, while precision and recall improved to 84.85% and 83.40%, respectively, resulting in a corresponding F1-score of 82.56%. These CLAHE-preprocessed images effectively increase local contrast in Pap smear images, making the morphological structures of cervical cells more distinguishable to the DenseNet-121 model. The improved contrast likely enables the network to learn more effectively and recognize cell features, such as size, texture, and boundary details—critical for distinguishing normal from abnormal cell types.

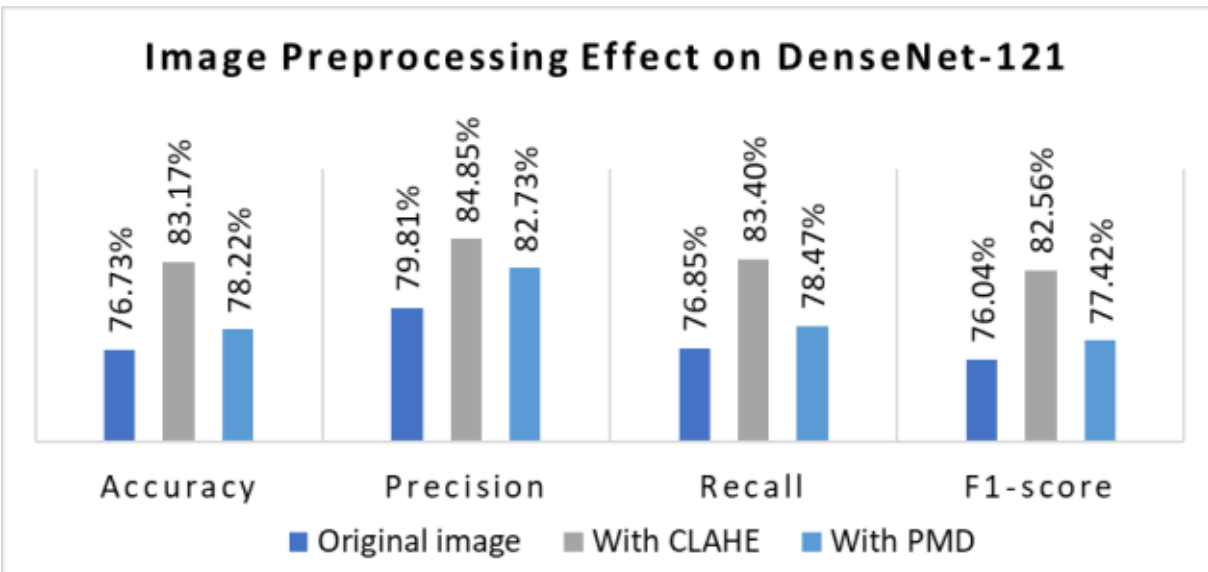

Figure 6. Effect of image preprocessing on DenseNet-121 performance

The preprocessing image with the PMD filter produced moderate improvements over the original dataset. The model achieved 78.22% accuracy, 82.73% precision, 78.47% recall, and an F1-score of 77.42%. These results suggest that PMD enhances image smoothness and edge preservation by reducing noise without blurring critical cell boundaries. However, it may not be as effective as CLAHE at improving overall contrast, which is necessary for fine-grained cellular classification. PMD still surpasses the raw dataset. Its effectiveness in pap smear image preprocessing. The comparative analysis demonstrates that CLAHE preprocessing consistently outperforms the original and PMD-preprocessed images. It reinforces its suitability for enhancing Pap smear images in deep learning applications. DenseNet-121, known for its dense connectivity and feature reuse, benefits significantly from CLAHE-enhanced image quality, resulting in superior classification performance.

**Discussion**

The experimental results demonstrate that image preprocessing is critical to improving CNN performance for Pap smear image classification. CLAHE consistently outperformed both the original images and PMD-preprocessed images. These results indicate that contrast enhancement contributes more significantly to classification performance than noise reduction. The qualitative observations in Figure 3 help explain this behavior. Pap smear classification relies heavily on subtle morphological characteristics, including nuclear shape, nuclear-cytoplasmic ratio, texture, and cell boundary information. In the original images, these features are often obscured by low contrast and uneven illumination. CLAHE enables CNN models to learn more representative and discriminative features, thereby improving classification performance. Similar findings have been reported in previous medical image analysis studies. The contrast enhancement improved feature extraction and classification accuracy.

The PMD filter focuses on noise suppression while preserving major structural boundaries. Although this process improves image smoothness and reduces unwanted variations. PMD filter does not substantially increase the contrast between important cellular components. Furthermore, excessive smoothing may attenuate fine textures and high-frequency details that are essential for distinguishing among cervical cell categories. PMD produced only moderate improvements for ResNet-34 and DenseNet-121 and slightly reduced the performance of MobileNet-V2. These findings suggest that preserving and enhancing diagnostically relevant morphological details is more beneficial for cervical cell classification than emphasizing noise reduction.

The influence of preprocessing also varied among the CNN architectures. ResNet-34 exhibited the largest improvement following CLAHE preprocessing, with accuracy increasing from 76.73% to 84.16%. It indicates that its residual learning mechanism effectively exploits enhanced image features. DenseNet-121 also benefited substantially from CLAHE due to its dense connectivity structure. DenseNet-121 promotes feature reuse and efficient information flow. MobileNet-V2 demonstrated a different behavior. Although it achieved the highest baseline accuracy among the evaluated models. MobileNet-V2 experienced only modest gains from CLAHE and a slight performance decline after PMD preprocessing. This behavior is attributed to its depthwise separable convolutions, which are more sensitive to the loss of fine-grained image details caused by PMD filtering. The findings indicate that cervical cell classification depends more strongly on enhancing morphological visibility than on aggressive noise suppression. A practical implication of this research is that a simple contrast enhancement technique can improve classification performance without increasing model complexity. This research is limited to a single dataset and two preprocessing techniques, which may affect the generalizability of the findings.

## CONCLUSION

This research investigated the relative effectiveness of contrast enhancement and noise reduction techniques for improving CNN-based cervical cancer classification on the SIPaKMeD dataset. Experimental results show that image preprocessing influences classification performance across all evaluated architectures (ResNet-34, DenseNet-121, and MobileNet-V2). CLAHE consistently achieved superior performance, increasing the classification accuracy of ResNet-34 from 76.73% to 84.16%, DenseNet-121 from 76.73% to 83.17%, and MobileNet-V2 from 83.91% to 84.65%. PMD filter provided only moderate improvements and slightly reduced the performance of MobileNet-V2. The findings indicate that contrast enhancement is more effective than noise reduction for improving feature representation and classification accuracy in Pap smear images. The main contribution of this research is a systematic comparison of CLAHE and PMD preprocessing techniques under identical experimental settings across three representative CNN architectures. This research is limited to a single dataset and two preprocessing methods. Future work will investigate optimized CLAHE parameters, hybrid preprocessing pipelines, and evaluation on additional cervical cytology datasets.

## ACKNOWLEDGEMENTS

The authors acknowledge the use of ChatGPT and Grammarly as writing assistance tools during the preparation of the manuscript. The authors bear full responsibility for the scientific content, methodology, data analysis, interpretation of results, and conclusions in this research.

## REFERENCES

[1] T. Wu, E. Lucas, F. Zhao, P. Basu, and Y. Qiao, “Artificial intelligence strengthens cervical cancer screening – present and future,” *Cancer Biol. Med.*, pp. 1–16, Sep. 2024, doi: 10.20892/j.issn.2095-3941.2024.0198.

[2] B. Chitra and S. S. Kumar, "Recent advancement in cervical cancer diagnosis for automated screening: a detailed review," *J. Ambient Intell. Humaniz. Comput.*, vol. 13, no. 1, pp. 251–269, Jan. 2022, doi: 10.1007/s12652-021-02899-2.

[3] S. L. Tan, G. Selvachandran, W. Ding, R. Paramesran, and K. Kotecha, "Cervical cancer classification from pap smear images using deep convolutional neural network models," *Interdiscip. Sci.*, vol. 16, no. 1, pp. 16–38, Mar. 2024, doi: 10.1007/s12539-023-00589-5.

[4] A. Khozaimi and W. Mahmudy F., "New insight in cervical cancer diagnosis using convolution neural network architecture," *IAES International Journal of Artificial Intelligence (IJ-AI)*, vol. 13, no. 3, p. 3092, Sep. 2024, doi: 10.11591/ijai.v13.i3.pp3092-3100.

[5] H. J. Hon, P. P. Chong, H. L. Choo, and P. P. Khine, "A Comprehensive Review of Cervical Cancer Screening Devices: The Pros and the Cons," *Asian Pacific Journal of Cancer Prevention*, vol. 24, no. 7, pp. 2207–2215, Jul. 2023, doi: 10.31557/APJCP.2023.24.7.2207.

[6] M. Li, Y. Jiang, Y. Zhang, and H. Zhu, "Medical image analysis using deep learning algorithms," *Front. Public Health*, vol. 11, Nov. 2023, doi: 10.3389/fpubh.2023.1273253.

[7] X. Kuang, F. Wang, K. M. Hernandez, Z. Zhang, and R. L. Grossman, "Accurate and rapid prediction of tuberculosis drug resistance from genome sequence data using traditional machine learning algorithms and CNN," *Sci. Rep.*, vol. 12, no. 1, p. 2427, Feb. 2022, doi: 10.1038/s41598-022-06449-4.

[8] C. Subaar *et al.*, "Investigating the detection of breast cancer with deep transfer learning using ResNet18 and ResNet34," *Biomed. Phys. Eng. Express*, vol. 10, no. 3, p. 035029, May 2024, doi: 10.1088/2057-1976/ad3cdf.

[9] N. Cinar, A. Ozcan, and M. Kaya, "A hybrid DenseNet121-UNet model for brain tumor segmentation from MR Images," *Biomed. Signal Process. Control*, vol. 76, p. 103647, Jul. 2022, doi: 10.1016/j.bspc.2022.103647.

[10] Q. Zhuang, S. Gan, and L. Zhang, "Human-computer interaction based health diagnostics using ResNet34 for tongue image classification," *Comput. Methods Programs Biomed.*, vol. 226, p. 107096, Nov. 2022, doi: 10.1016/j.cmpb.2022.107096.

[11] A. Souid, N. Sakli, and H. Sakli, "Classification and Predictions of Lung Diseases from Chest X-rays Using MobileNet V2," *Applied Sciences*, vol. 11, no. 6, p. 2751, Mar. 2021, doi: 10.3390/app11062751.

[12] T. Sood, P. Khandnor, and R. Bhatia, "Enhancing pap smear image classification: integrating transfer learning and attention mechanisms for improved detection of cervical abnormalities," *Biomed. Phys. Eng. Express*, vol. 10, no. 6, p. 065031, Nov. 2024, doi: 10.1088/2057-1976/ad7bc0.

[13] Y. Karasu Benyes, E. C. Welch, A. Singhal, J. Ou, and A. Tripathi, "A Comparative Analysis of Deep Learning Models for Automated Cross-Preparation Diagnosis of Multi-Cell Liquid Pap Smear Images," *Diagnostics*, vol. 12, no. 8, p. 1838, Jul. 2022, doi: 10.3390/diagnostics12081838.

[14] L. Chato, K. T. Kashyap, K. P. Marupaka, and K. Santosh, "Hybrid DL Classification Model Based on CNN and Transformer for Pap Smear Cells," in *2025 IEEE 22nd International Symposium on Biomedical Imaging (ISBI)*, IEEE, Apr. 2025, pp. 1–4. doi: 10.1109/ISBI60581.2025.10981236.

[15] B. J. Chelliah, S. K. Gahra, M. Shrinidhi, S. Y. Devi, A. SenthilSelvi, and P. Meenalohchini, "Enhancing Cervical Cancer Detection: Explainable AI and Attention Mechanisms for Pap Smear Classification," in *2025 8th International Conference on Circuit, Power & Computing Technologies (ICCPCT)*, IEEE, Aug. 2025, pp. 1667–1672. doi: 10.1109/ICCPCT65132.2025.11176745.

[16] H. Alquran, M. Alsalatie, W. A. Mustafa, R. Al Abdi, and A. R. Ismail, "Cervical Net: A Novel Cervical Cancer Classification Using Feature Fusion," *Bioengineering*, vol. 9, no. 10, p. 578, Oct. 2022, doi: 10.3390/bioengineering9100578.

[17] J. Gangrade, R. Kuthiala, S. Gangrade, Y. P. Singh, M. R, and S. Solanki, "A deep ensemble learning approach for squamous cell classification in cervical cancer," *Sci. Rep.*, vol. 15, no. 1, p. 7266, Mar. 2025, doi: 10.1038/s41598-025-91786-3.

[18] S. Dodge and L. Karam, "Understanding how image quality affects deep neural networks," in *2016 Eighth International Conference on Quality of Multimedia Experience (QoMEX)*, IEEE, Jun. 2016, pp. 1–6. doi: 10.1109/QoMEX.2016.7498955.

[19] K. P. Win, Y. Kitjaidure, K. Hamamoto, and T. Myo Aung, "Computer-assisted screening for cervical cancer using digital image processing of pap smear images," *Applied Sciences*, vol. 10, no. 5, p. 1800, Mar. 2020, doi: 10.3390/app10051800.

[20] S. A. Al-asbaily, S. Almoshity, S. Younus, and K. Bozed, "Classification of Cervical Cancer using Convolutional Neural Networks," in *2024 IEEE 4th International Maghreb Meeting of the

*Conference on Sciences and Techniques of Automatic Control and Computer Engineering (MI-STA)*, IEEE, May 2024, pp. 735–739. doi: 10.1109/MI-STA61267.2024.10599701.
[21] V. Malathi and A. Manikandan, "An underwater image enhancement by reducing speckle noise using modified anisotropic diffusion filter," *International Journal of Electrical and Computer Engineering (IJECE)*, vol. 13, no. 6, p. 6361, Dec. 2023, doi: 10.11591/ijece.v13i6.pp6361-6368.
[22] E. A. Tjoa, I. P. Yowan Nugraha Suparta, R. Magdalena, and N. Kumalasari CP, "The use of CLAHE for improving an accuracy of CNN architecture for detecting pneumonia," *SHS Web of Conferences*, vol. 139, p. 03026, May 2022, doi: 10.1051/shsconf/202213903026.
[23] M. E. Plissiti, P. Dimitrakopoulos, G. Sfikas, C. Nikou, O. Krikoni, and A. Charchanti, "Sipakmed: A New Dataset for Feature and Image Based Classification of Normal and Pathological Cervical Cells in Pap Smear Images," in *2018 25th IEEE International Conference on Image Processing (ICIP)*, 2018, pp. 3144–3148. doi: 10.1109/ICIP.2018.8451588.
[24] B. Z. Wubineh, A. Rusiecki, and K. Halawa, "Segmentation and Classification Techniques for Pap Smear Images in Detecting Cervical Cancer: A Systematic Review," *IEEE Access*, vol. 12, pp. 118195–118213, 2024, doi: 10.1109/ACCESS.2024.3447887.
[25] C. Avatavului and M. Prodan, "Evaluating Image Contrast: A Comprehensive Review and Comparison of Metrics," *Journal of Information Systems & Operations Management*, vol. 17, no. 1, pp. 143–160, 2023.
[26] K. G. Dhal, A. Das, S. Ray, J. Gálvez, and S. Das, "Histogram equalization variants as optimization problems: a review," *Archives of Computational Methods in Engineering*, vol. 28, no. 3, pp. 1471–1496, May 2021, doi: 10.1007/s11831-020-09425-1.
[27] J. C. M. dos Santos, G. A. Carrijo, C. de Fátima dos Santos Cardoso, J. C. Ferreira, P. M. Sousa, and A. C. Patrocínio, "Fundus image quality enhancement for blood vessel detection via a neural network using CLAHE and Wiener filter," *Research on Biomedical Engineering*, vol. 36, no. 2, pp. 107–119, Jun. 2020, doi: 10.1007/s42600-020-00046-y.
[28] Y. R. Haddadi, B. Mansouri, and F. Z. I. Khodja, "A novel medical image enhancement algorithm based on CLAHE and pelican optimization," *Multimed. Tools Appl.*, Apr. 2024, doi: 10.1007/s11042-024-19070-6.
[29] P. Perona and J. Malik, "Scale-space and edge detection using anisotropic diffusion," *IEEE Trans. Pattern Anal. Mach. Intell.*, vol. 12, no. 7, pp. 629–639, Jul. 1990, doi: 10.1109/34.56205.
[30] A. Khozaimi, I. Darti, S. Anam, and W. M. Kusumawinahyu, "Advanced cervical cancer classification: enhancing pap smear images with hybrid PMD filter-CLAHE," *Indonesian Journal of Electrical Engineering and Computer Science*, vol. 39, no. 1, p. 644, Jul. 2025, doi: 10.11591/ijeecs.v39.i1.pp644-655.
[31] Q. Li, Y. Hu, and Y. Cao, "Adaptive Perona–Malik model based on dynamical threshold for image multi–noise removal with details preservation," *Computers & Mathematics with Applications*, vol. 137, pp. 28–43, May 2023, doi: 10.1016/j.camwa.2023.02.012.
[32] O. Attallah, "CerCan·Net: Cervical cancer classification model via multi-layer feature ensembles of lightweight CNNs and transfer learning," *Expert Syst. Appl.*, vol. 229, p. 120624, Nov. 2023, doi: 10.1016/j.eswa.2023.120624.
[33] S. Goyal, P. Thakur, and A. Laddi, "Gaze estimation optimization through performance analysis of ResNet variants," *Procedia Comput. Sci.*, vol. 260, pp. 880–887, 2025, doi: 10.1016/j.procs.2025.03.270.
[34] H. Yuan, J. Cheng, Y. Wu, and Z. Zeng, "Low-res MobileNet: An efficient lightweight network for low-resolution image classification in resource-constrained scenarios," *Multimed. Tools Appl.*, vol. 81, no. 27, pp. 38513–38530, Nov. 2022, doi: 10.1007/s11042-022-13157-8.
[35] S. Bhoopal, M. Rao, and C. H. Krishnappa, "Enhanced diabetic retinopathy detection and classification using fundus images with ResNet50 and CLAHE-GAN," *Indonesian Journal of Electrical Engineering and Computer Science*, vol. 35, no. 1, p. 366, Jul. 2024, doi: 10.11591/ijeecs.v35.i1.pp366-377.
[36] A. R. Breitner and Y. Bandung, "Calculation of Museum Collections Popularity Using Detection and Tracking of Visitors From Surveillance Cameras in Low Light Conditions," in *2024 International Seminar on Intelligent Technology and Its Applications (ISITIA)*, IEEE, Jul. 2024, pp. 316–321. doi: 10.1109/ISITIA63062.2024.10667676.
[37] M. Heydarian, T. E. Doyle, and R. Samavi, "MLCM: Multi-Label Confusion Matrix," *IEEE Access*, vol. 10, pp. 19083–19095, 2022, doi: 10.1109/ACCESS.2022.3151048.